\documentclass{article}
\usepackage{arxiv}
\usepackage{array}
\usepackage{tabularx}
\usepackage{makecell}
\usepackage{threeparttable}
\usepackage{amsmath,amssymb,amsfonts}
\usepackage{graphicx}
\usepackage{graphics}
\usepackage{booktabs}
\usepackage{CJKutf8}  
\usepackage{bbding}   
\usepackage{algorithm}  
\usepackage{algorithmicx}
\usepackage{algpseudocode}
\usepackage{color}
\usepackage{multirow}
\usepackage{booktabs}
\usepackage{hyperref}
\usepackage{float}
\usepackage{subfig}
\usepackage{textcomp}
\usepackage[utf8]{inputenc} 
\usepackage[T1]{fontenc}    
\usepackage{hyperref}       
\usepackage{url}            
\usepackage{booktabs}       
\usepackage{amsfonts}       
\usepackage{nicefrac}       
\usepackage{microtype}      
\usepackage{lipsum}
\usepackage{balance}
\usepackage{stfloats}
\usepackage{url}
\usepackage{verbatim}
\usepackage{graphicx}
\graphicspath{ {./images/} }

\algnewcommand\algorithmicinput{\textbf{Input: }}
\algnewcommand\Input{\item[\algorithmicinput]}
\algnewcommand\algorithmicoutput{\textbf{Output: }}
\algnewcommand\Output{\item[\algorithmicoutput]}
\def\tsc#1{\csdef{#1}{\textsc{\lowercase{#1}}\xspace}}
\tsc{WGM}
\tsc{QE}
\title{Safe Navigation for Robotic Digestive Endoscopy via Human Intervention-based Reinforcement Learning}

\author{
 Min Tan \\
  Shenzhen Institutes of Advanced Technology\\ Chinese Academy of Sciences \\
  Shenzhen 518055, China \\
  University of Chinese Academy of Sciences \\
  Beijing 101400, China \\
  \texttt{min.tan@siat.ac.cn} \\
  \And
 Yushun Tao \\
  Shenzhen Institutes of Advanced Technology\\Chinese Academy of Sciences \\
  Shenzhen 518055, China \\
  University of Chinese Academy of Sciences \\
  Beijing 101400, China \\
  \texttt{ys.tao@siat.ac.cn} \\
  \And
 Boyun Zheng \\
  Department of Electronic Engineering\\ Chinese University of Hong Kong \\
  Hong Kong SAR 999077, China \\
  \texttt{byzheng@link.cuhk.edu.hk} \\
  \And
 Gaosheng Xie \\
  Shenzhen Institutes of Advanced Technology\\ Chinese Academy of Sciences \\
  Shenzhen 518055, China \\
  \texttt{gs.xie@siat.ac.cn} \\
  \And
 Lijuan Feng \\
  Department of Gastroenterology and Hepatology\\Shenzhen University General Hospital \\
  Shenzhen 518055, China \\
  \texttt{fenglj@szu.edu.cn} \\
  \And
 Zeyang Xia \\
  School of Mechanical Engineering\\ Shanghai Jiao Tong University \\
  Shanghai 200240, China \\
  \texttt{zxia@sjtu.edu.cn} \\
  \And
 Jing Xiong \\
  Shenzhen Institutes of Advanced Technology\\ Chinese Academy of Sciences \\
  Shenzhen 518055, China \\
  University of Chinese Academy of Sciences \\
  Beijing 101400, China \\
  \texttt{jing.xiong@siat.ac.cn} \\
}

\begin{document}
\maketitle
\begin{abstract}
With the increasing application of automated robotic digestive endoscopy (RDE), ensuring safe and efficient navigation in the unstructured and narrow digestive tract has become a critical challenge. Existing automated reinforcement learning navigation algorithms often result in potentially risky collisions due to the absence of essential human intervention, which significantly limits the safety and effectiveness of RDE in actual clinical practice. To address this limitation, we proposed a Human Intervention (HI)-based Proximal Policy Optimization (PPO) framework, dubbed HI-PPO, which incorporates expert knowledge to enhance RDE's safety. Specifically, HI-PPO combines Enhanced Exploration Mechanism (EEM), Reward-Penalty Adjustment (RPA), and Behavior Cloning Similarity (BCS) to address PPO's exploration inefficiencies for safe navigation in complex gastrointestinal environments. Comparative experiments were conducted on a simulation platform, and the results showed that HI-PPO achieved a mean ATE (Average Trajectory Error) of \(8.02\ \text{mm}\) and a Security Score of \(0.862\), demonstrating performance comparable to human experts. The code will be publicly available once this paper is published. \href{https://tokymin.github.io/hirde.index}{http://siat-medical-robot-hippo.index}.
\end{abstract}


\section{Introduction}
Clinical digestive endoscopy examinations are crucial medical procedures, regarded as the gold standard in early diagnosis of colorectal cancer \cite{chen_chen_toll_2017, martin_enabling_2020}. Traditional endoscopy carries the risk of endoscopic looping and, more seriously, a perforation rate of 0.1\% to 0.3\% during insertion \cite{ciuti_frontiers_2016, manfredi_endorobots_2021, huang_autonomous_2021, pore_autonomous_2023}. These complications can cause patient pain and potential tissue damage \cite{rex_quality_2015, chen_endoscopist_2007}.
In contrast, automated Robotic Digestive Endoscopy (RDE), including magnetic endoscope robots \cite{martin_enabling_2020, huang_autonomous_2021} and vision-based autonomous navigation robots \cite{mahmoud_orbslam-based_2017, prendergast_autonomous_2018} are designed to overcome the problems of manually navigating the endoscope, such as the non-intuitive control and the perforation risk. 
However, as the use of RDE systems increases, achieving safe and efficient navigation through the unstructured and narrow digestive tract has become a critical challenge. 
\begin{figure}
    \centering
    \includegraphics[width=1\linewidth]{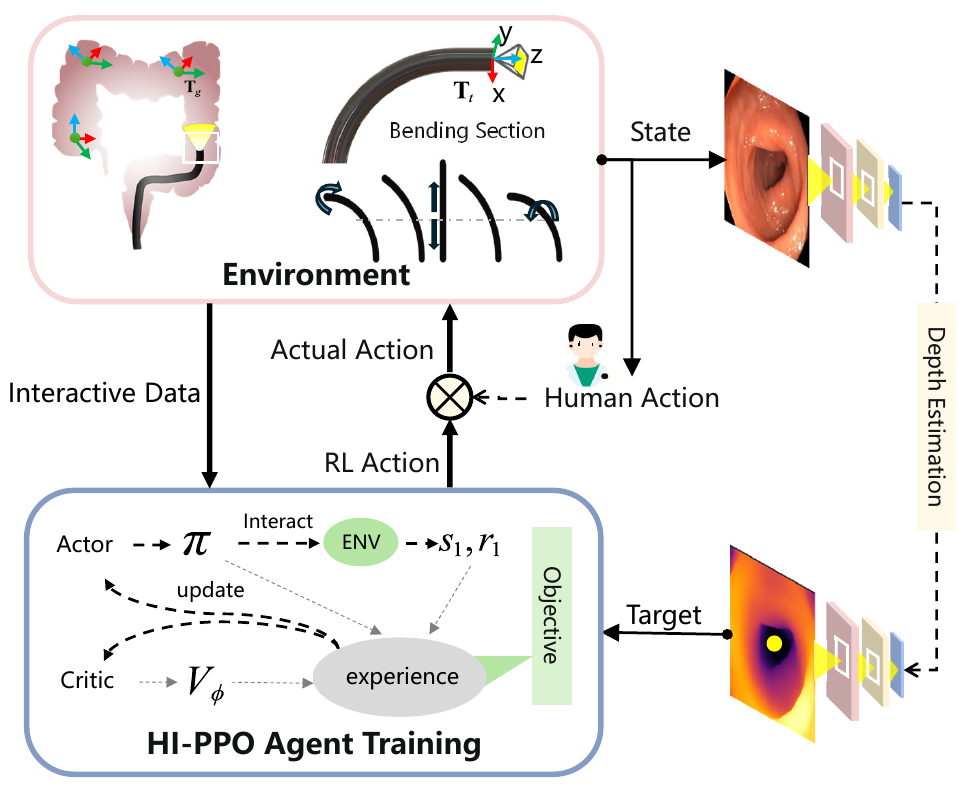}
    \caption{Overview framework of the proposed HI-PPO.
    }
    \label{fig:overview}
\end{figure}
Currently, several medical endoscopic robots have adopted reinforcement learning (RL) for navigation \cite{trovato2010development, corsi_constrained_2023, pore_colonoscopy_2022, lazo_autonomous_2022, zhang2022deep}. This approach excels in learning complex control strategies through environmental interactions, making it well-suited for dynamically changing and irregularly structured environments like human body lumens. Relevant to our work, Davide et al. \cite{corsi_constrained_2023} introduced Constrained Proximal Policy Optimization (C-PPO). By incorporating formal verification techniques, this method reduces the likelihood of agents entering extreme regions, thereby minimizing risks such as tissue damage or perforation. However, C-PPO's reliance on four discretized actions paired with sparse image blocks neglects continuous steering motions, introducing safety risks that limit its clinical applicability.

Automated RL-based navigation algorithms lacking human intervention (HI) may lead to unsafe collisions \cite{corsi_constrained_2023, ng_navigation_2024}. To address this, recent studies \cite{Wu2023HumanGuidedRL, saunders2018trial} have explored integrating HI into RL frameworks. For instance, Ameya et al. \cite{pore_colonoscopy_2022} developed a vision-based RL control system that uses human-in-the-loop supervision to replace risky agent actions with safe human demonstrations during emergencies. A more effective strategy involves imitating human behavior. Vecerik et al. \cite{vecerik2017leveraging} combined Behavioral Cloning (BC) with Deep Deterministic Policy Gradient (DDPG), outperforming conventional RL approaches by incorporating human demonstration data.

Despite these advancements, HI-augmented RL navigation faces two critical challenges: 1) RL systems relying solely on post-training human supervision fail to utilize prior knowledge, leading to high training costs; 2) Existing HI methods focus on isolated algorithm components rather than providing a unified integration framework, hindering safe and efficient navigation in narrow and irregular digestive tracts.

To address these gaps, we propose HI-PPO, a Human Intervention-enhanced Proximal Policy Optimization framework. HI-PPO improves robotic digestive endoscopy safety and efficiency by 1) Using behavioral cloning to mimic expert actions during training and 2) Incorporating HI to enhance exploration efficiency. Unlike prior work \cite{turan2019learning, pore_colonoscopy_2022, corsi_constrained_2023}, our method models the distal bending section of commercial colonoscopes, accounting for complex turning motions. Navigation targets are derived from pre-trained network depth estimates of endoscopic images.

Our main contributions are as follows:
\begin{itemize}
    \item First modeling of commercial colonoscope distal bending sections using HI-augmented RL, improving RDE safety through expert knowledge integration.
    \item A unified framework combining Enhanced Exploration Mechanism (EEM), Reward-Penalty Adjustment (RPA), and Behavior Cloning Similarity (BCS) to address PPO's exploration inefficiencies for safe navigation in complex gastrointestinal environments.
    \item Simulation-based validation demonstrating HI-PPO's superiority over existing methods, achieving human-expert-level performance.
\end{itemize}

\section{Related work}
\subsection{RL-Based Autonomous Navigation for Endoscopes}
Current research widely employs model-free deep reinforcement learning (DRL) to address navigation control challenges for endoscopes in complex luminal environments. For instance, Ng et al.\cite{ng_navigation_2024} proposed a Proximal Policy Optimization (PPO)-based control method for tendon-driven flexible endoscopes, achieving over 90\% navigation success rate with a 3 mm tolerance through simulation training, while demonstrating the transfer capability of pre-trained policies in contact environments. Trovato et al.\cite{trovato2010development} implemented early autonomous control for colonoscopy robots using Q-learning algorithms, albeit limited by discrete action space design. Zhang et al.\cite{zhang2022deep} and Turan et al.\cite{turan2019learning} applied PPO and DDPG algorithms to capsule endoscope control, attaining millimeter-level tracking accuracy in gastric coverage scanning and magnetic navigation tasks. Expanding beyond gastrointestinal applications, Li et al.\cite{li_rl-tee_2023} developed an attention-augmented RL framework (RL-TEE) for ultrasound probe navigation, achieving compliant 3-DOF control in deformable esophageal environments through hybrid CNN-self-attention architectures. However, these methods generally suffer from oversimplified action space designs, exemplified by the C-PPO method\cite{corsi_constrained_2023}, which encodes only four discrete actions while neglecting continuous steering motion modeling, resulting in safety risks in practical applications.

\subsection{Human Intelligence-Integrated Reinforcement Learning}
To enhance navigation safety and training efficiency, researchers have explored integrating human intervention (HI) mechanisms into RL frameworks. Wu et al.\cite{Wu2023HumanGuidedRL} developed a human-robot collaborative RL framework that achieves safe navigation in dynamic environments through prioritized experience replay and reward shaping mechanisms during sim-to-real transfer. Pore et al.\cite{pore_autonomous_2023} and Lazo et al.\cite{lazo_autonomous_2022} employed CNN-based visual servoing systems with ergonomic soft robot designs, improving model generalization through cross-domain data training. Vecerik et al.\cite{vecerik2017leveraging} innovatively combined behavior cloning (BC) with DDPG, utilizing human demonstrations to address sparse reward problems, showing significant performance improvements over conventional RL algorithms. Nevertheless, current approaches still face limitations: Saunders et al.\cite{saunders2018trial} require continuous human monitoring for online intervention, incurring high training costs; Wang et al.\cite{wang2018intervention}'s intervention-aided RL reduces human involvement frequency but lacks a systematic framework for integrating prior knowledge, hindering adaptation to complex geometric constraints in narrow digestive tract lumens.  

While autonomous control remains a key objective in endoscopic robotics, medical safety demands necessitate human intervention integration beyond mere post-training supervision. Current research trends embedded human expertise via behavior cloning and guided policy exploration. However, systematically integrating human priors during training while maintaining endoscopic safety constraints remains a critical challenge.

\section{Methodology}
This study aims to develop a robotized flexible endoscope autonomous navigation system based on deep reinforcement learning (DRL). As described in Section 3.1, we first constructed a reinforcement learning problem model incorporating a hybrid discrete-continuous action space. Section 3.2 elaborates on the triple-enhancement mechanism integrating human intervention-enhancing strategy safety and learning efficiency through Expert-guided Exploration (EEM), Reward-Penalty Adjustment (RPA), and Behavior Cloning Strategy (BCS). Finally, Section 3.3 proposes the H-PPO integration framework, which combines prioritized experience replay and safety coverage mechanisms to achieve optimized navigation strategies with human intervention.
\subsection{DRL Problem Formulation for RDE Agent}
\textbf{Observation Space:}
The network input of the agent consists of 128×128-pixel RGB images, where visual features are extracted through a CNN encoder. Depth maps are independently generated by a pre-trained Depth Anything network\cite{yang2024depth}, exclusively used for computing the navigation target point \(p_{target}\). The target coordinates \((x,y)\) and its relative distance \(d_{target}\) to the endoscope are encoded into a 4-dimensional vector (normalized to \([-1,1]\)), which is concatenated with the 128-dimensional visual features from the CNN, forming a 132-dimensional State representation. This design effectively reduces computational overhead while ensuring the transmission of critical geometric information required for spatial reasoning to the policy network.

\textbf{Action Space:}
The agent's action space adopts a hybrid discrete-continuous architecture, comprising two independent decision branches for motion control and bending control, achieving 6-DOF manipulation through the Unity physics engine. The motion control branch outputs a 3D continuous action vector \(\mathbf{v}\in[-1,1]^3\), which is directly converted into instantaneous displacement increments \(\Delta \mathbf{p} = \mathbf{v} \cdot 10\text{mm/s} \cdot 0.1s\) at the endoscope tip in the virtual environment. This displacement is constrained by depth map-based lumen visibility analysis—when the area of the largest connected region falls below 5\% of $128^2$ pixels (i.e., \(L=0\)), safety mechanisms override speed control. The bending control branch employs a quaternary discrete action set \(\{0:\text{forward bend}, 1:\text{backward bend}, 2:\text{left bend}, 3:\text{right bend}\}\), where each action triggers a \(5^\circ\) bending angle increment. The Unity-built-in joint constraint system ensures bending ranges do not exceed \(\pm90^\circ\) in each direction. When angles exceed limits, the physics engine automatically executes progressive resetting: \(\theta_j(t+1) = \theta_j(t) - \text{sgn}(\theta_j(t)) \cdot 5^\circ/\text{frame}\) until safe posture is restored. Action execution follows a mode-exclusive principle: translation DOFs are locked via an animation State machine during bending to prevent non-physical deformation of virtual soft tissues caused by composite motions.

\begin{figure}
    \centering
    \includegraphics[width=1\linewidth]{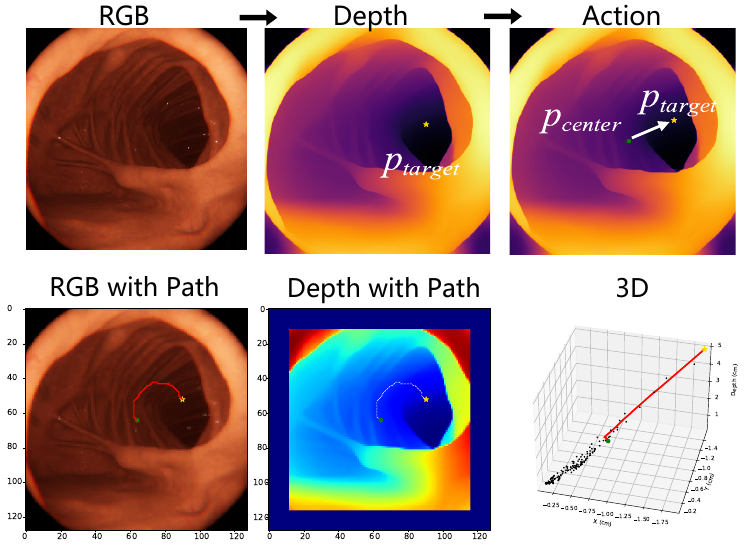}
    \caption{Detection of endoscopic navigation target point using depth estimation and connected component centroid analysis.
    }
    \label{fig:depth_action}
\end{figure}

\textbf{Reward Function:}
Our primary objective is to make the RDE agent reach the colon terminus while avoiding tissue damage. To achieve this, the agent must maintain the depth map-derived local navigation point \(p_{target}\) near the image center \(p_{center}=(64,64)\). We construct a dense reward function based on this principle:  
\begin{equation}
r_{t} = \begin{cases}
\left(1 - \frac{||p_{target} - p_c||_2}{64}\right), & \text{target visible }(L=1) \\
-0.5, & \text{target lost }(L=0)
\end{cases}
\end{equation}

 where the normalization factor \(D_{max}=64\) corresponds to half the image width. The visibility flag \(L\) is determined in real-time through connected-component analysis of the depth map—valid when the largest connected region exceeds 5\% of $128^2$ pixels. To detect \(p_{target}\) in endoscopic images, we implement a depth estimation algorithm based on \cite{yang2024depth} (see Figure \ref{fig:depth_action}), where \(p_{target}\) is computed as the centroid of the farthest connected domain from the endoscope in the depth map. A terminal reward of +10 is granted upon successful navigation to the endpoint, while a penalty of -10 is applied if the agent returns to the starting point without human intervention.

\subsection{Human Intervention Techniques}
To enhance the safety of RDE navigation, we propose three mechanisms to enable real-time navigation actions to adapt to complex environments. Enhanced Exploration Mechanism (EEM) integrates expert knowledge by improving State exploration efficiency, where human intervention takes control when agents get stuck. The Reward-Penalty Adjustment (RPA) promotes safer policy learning by penalizing unsafe actions through a penalty score during initial interventions. Behavior Cloning Similarity (BCS) improves learning performance by guiding the agent to imitate expert actions, integrating similarity objectives into the actor-critic network for optimal policy learning. 

\textbf{Enhanced Exploration Mechanism (EEM):}
While standard PPO expands the agent's exploration space through stochasticity, unsafe exploration in confined colonoscopy scenarios may increase patient injury risks. Specifically, the high-dimensional State space and complex navigation paths in colonic environments require extensive samples due to greedy exploration, leading to prolonged training times and suboptimal trajectories \cite{corsi_constrained_2023}.  

In contrast, human-guided exploration enhances this process through real-time intervention and expert knowledge, replacing inefficient actions with expert-corrected ones. We propose the Enhanced Exploration Mechanism (EEM) that modifies the behavior policy \(\pi^{\text{b}}\) to incorporate human actions during interventions:  
\begin{equation}
\pi^{\text{b}}(a_t | s_t) = \begin{cases} 
  \pi_\theta(a_t|s_t), & M_t=0 \\
  \delta(a_t=a_t^H), & M_t=1 
  \end{cases}
\end{equation}

where \(M_t\) is a binary intervention indicator (\(M_t=1\) during human intervention; otherwise \(M_t=0\)), and \(a_t^{\text{H}}\) denotes the human action at timestep \(t\). This mechanism ensures direct integration of expert demonstrations into the learning process, improving data quality and exploration efficiency.

\textbf{Reward-Penalty Adjustment (RPA):} An instantaneous intervention penalty term is added to the base reward \(r_t\):  
\begin{equation}
r_t^{\text{new}} = r_t + \lambda \cdot \delta(M_t=1 \land M_{t-1}=0)
\end{equation}

where \(\lambda = -2.5\) quantifies the intervention penalty magnitude, and the penalty is triggered exclusively when human intervention is initiated (\(M_t=1\) with \(M_{t-1}=0\)). This design strategically applies a substantial negative reward at the moment of intervention activation, which serves two synergistic purposes: it incentivizes the agent to proactively avoid behaviors that necessitate human takeover, while maintaining policy optimization stability by restricting penalty imposition to discrete intervention onset events rather than continuous penalization throughout the intervention period.

\textbf{Behavior Cloning Strategy (BCS):}
In addition to reward adjustment, we enhance policy learning by using behavior cloning as an auxiliary objective. Specifically, we add an action similarity term to the standard PPO objective function, enabling the agent to imitate expert behaviors while optimizing the cumulative reward. The BCS-integrated objective function can be denoted as follows:
\begin{equation}
\label{init_objective}
\mathcal{J}(\theta)=\mathbb{E}_t\left[\min\left(r_t(\theta)A_t,\text{clip}\left(r_t(\theta),1 - \epsilon,1+\epsilon\right)A_t\right)\right]\end{equation}
\[+\alpha\cdot\text{sim}(a_t^\theta,a_t^{\text{H}})\]
where \(\text{sim}(\cdot)\) is the similarity metric, and \(\alpha\in[0,1]\) is the weight coefficient for imitation learning. This design directly aligns the agent's action output with the deterministic actions of human experts on a per-dimension basis, avoiding the incompatibility problem between probability distribution assumptions and deterministic actions. 
\begin{algorithm}[tbp]
\caption{H-PPO of RDE}
\begin{algorithmic}[1]
\Input{Environment $Env$, human policy $\pi^H$, training epochs $E$}
\Output{Optimized policy $\pi_\theta$}
\State Estimate depth map \(\mathbf{D}_t \leftarrow \text{DepthAnything}(\mathbf{I}_t)\)
\State Initialize $\pi_\theta$, $V_\theta$, replay buffer $D$, priority coefficient $\beta$, safety timer $\tau=0$
\For{episode = 1 to $E$}
    \State Reset $Env$, obtain initial State $s_1$, initialize trajectory buffer $Q=\emptyset$
    \While{not terminated}
        \If{human intervenes}
            \State $a_t \leftarrow \pi^H(s_t)$, $M_t\leftarrow1$
        \Else
            \State $a_t \leftarrow \pi_\theta(s_t)$, $M_t\leftarrow0$
        \EndIf
        \State \textbf{Safety Override:}
        \If{target lost ($L=0$)}
            \State $a_t[vel] \leftarrow 0$ Freeze insertion speed 
            \State $\tau \leftarrow \tau + \Delta t$
            \If{$\tau \geq 5s$}
                \State Extract recent 50 positions from $Q$, compute smoothed path $\{\tilde{q}_k\}$
                \State $a_t[vel] \leftarrow -0.5 \cdot \text{sat}(||q_t - \tilde{q}_{t-1}||_2/10\text{mm})$
            \EndIf
        \Else
            \State $\tau \leftarrow 0$
        \EndIf
        \State Execute $a_t$, observe $r_t$, $s_{t+1}$, append pre-action position $q_t^{pre}$ to $Q$
        \State Compute modified reward $r_t' = r_t -2.5 \cdot \delta(M_t=1)$
        \State Store transition $\zeta_t = (s_t,a_t,M_t,r_t',s_{t+1})$ in $D$
        \State Sample batch, update priorities via $\hat{\rho}_i = |V_\theta(s_i)-V_{target}| + \beta\cdot\max(0, A^H_i - A^{RL}_i)$
    \EndWhile
    \State Update $V_\theta$ and $\pi_\theta$
\EndFor
\end{algorithmic}
\label{algorithm_motion}
\end{algorithm}
\subsection{H-PPO Framework Integration}
\textbf{Prioritized Experience Replay Optimization:}
Experience sampling probabilities are determined by \(p(\varsigma_i) \propto \hat{\rho}_i^\alpha\). This mechanism prioritizes experiences with high critic prediction errors and significant human-RL advantage discrepancies during multiple PPO update epochs, thereby enhancing expert experience utilization. Within the PPO framework, experience priority \(\hat{\rho}_i\) integrates critic network errors and human action advantages:  
\begin{equation}
\hat{\rho}_i = \underbrace{|V_\theta(s_i) - V_{target}|}_{\text{Critic error}} + \beta \cdot \underbrace{\max(0, A^H_i - A^{RL}_i)}_{\text{Human-RL advantage gap}} + \epsilon
\end{equation}

where \(V_{target} = \sum_{k=0}^{n-1} \gamma^k r_{t+k} + \gamma^n V_\theta(s_{t+n})\) denotes n-step returns, \(\epsilon \in R^{+}\) is a small normal number to ensure that the value is greater than zero, and \(A^H_i\) represents 
human action advantage computed via actual trajectory returns:  
  \begin{equation}
  A^H_i = \sum_{k=0}^{T-t} \gamma^k r_{t+k}^H - V_\theta(s_t)
  \end{equation}
where \(A^{RL}_i\) estimates RL action advantage using Generalized Advantage Estimation (GAE):  
  \begin{equation}
  A^{RL}_i = \sum_{l=0}^{T-t} (\gamma\lambda)^l \delta_{t+l}, \quad \delta_t = r_t + \gamma V_\theta(s_{t+1}) - V_\theta(s_t)
 \end{equation}
where \(\beta\) modulates human guidance intensity, and \(\epsilon\) ensures minimal sampling probability for all experiences.

As a result, the retrieving probability of the tuple \(\varsigma_i\) is calculated by the following probability density function \(p^{new}\):
\begin{equation} p(i)^{new} = \frac{\hat{\rho}_i^{\alpha}}{\sum_{k = 1}^{|D|} \hat{\rho}_k^{\alpha}} \end{equation}
where \(\alpha\) is the scaling coefficient and \(|D|\) is the size of the experience replay buffer.

\textbf{Imitation Learning Enhancement:}
Building upon PPO's clipping mechanism, this design incorporates a KL divergence constraint to align policy updates with expert behavioral patterns while maximizing expected returns. The policy optimization objective is enhanced through human demonstration guidance. The modified policy loss function from Equation \ref{init_objective} becomes:  
\begin{equation}
\mathcal{J}(\theta)= \mathbb{E}_t\left[\min(r_t(\theta)A_t, \text{clip}(r_t(\theta),1-\epsilon,1+\epsilon)A_t)\right] \label{imitation}\end{equation}
\[+ \lambda \cdot \omega_t \cdot D_{KL}(\pi_\theta(\cdot|s_t) \| \pi^H(\cdot|s_t))\]

where \(r_t(\theta) = \frac{\pi_\theta(a_t|s_t)}{\pi_{\theta_{old}}(a_t|s_t)}\) denotes the policy probability ratio, \(A_t\) represents the Generalized Advantage Estimation (GAE), and \(D_{KL}\) quantifies the Kullback-Leibler divergence between current and human policies. The adaptive weight \(\omega_t = \frac{\max(0, A^H_t - A^{RL}_t)}{\sum A^H_t}\) dynamically scales the KL term, where \(A^H_t\) is the human action advantage value.  Equation \ref{imitation} ensures progressive policy alignment with expert behaviors while maintaining PPO's stable optimization characteristics.

\textbf{Security mechanism integration:}
\label{Securitymechanismintegration}
The algorithm interleaves RL and human actions per episode, storing experiences with intervention markers. Prioritized replay elevates critical sample usage by merging value errors and human-RL advantage gaps. At the same time, policy updates blend PPO objectives with behavior cloning loss for efficient human-RL co-learning. A safety protocol operates to mitigate depth calculate failures: upon target loss (\(L=0\)), forward velocity is halted (\(a_t[vel] \leftarrow 0\)) with bending retained for reorientation; if vision remains lost $>$ 5 s, a smooth retraction path is generated from the last 50 trajectory points, guiding withdrawal at $\leq$ 0.5 mm/s until visibility recovers or safe zones are reached. Path smoothing and velocity saturation ensure tissue protection during the emergency retreats.
\begin{figure}[h]
    \centering
    \vfill
\includegraphics[width=1\linewidth]{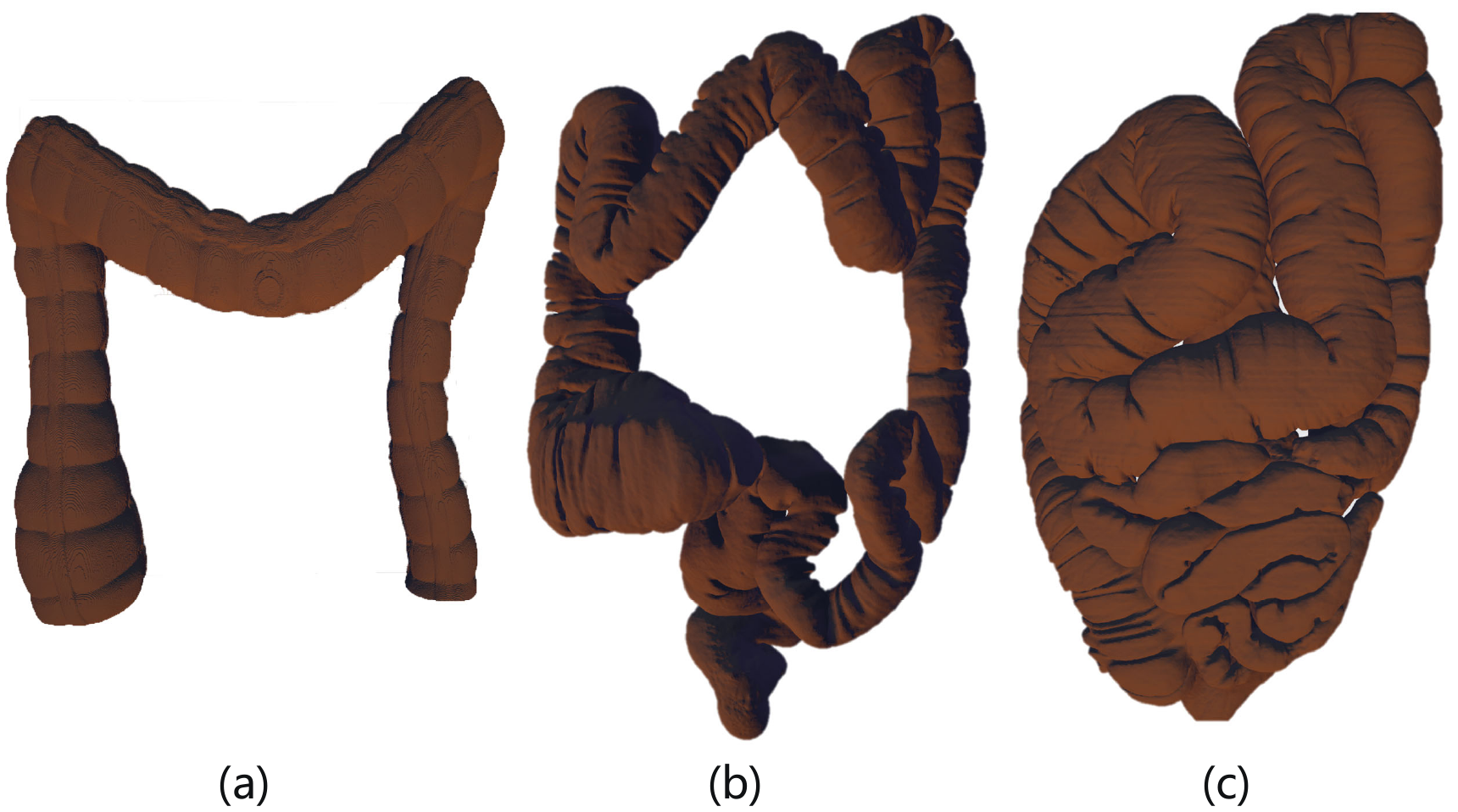}
    \caption{The colon model used in the experimental phase. (a) Simple colon No.1 for training (b) Complex colon No.2 for training (c) Complex colon No.3 for inference.}
    \label{fig:colon_model}
\end{figure}

\section{Experimental Verification}
This study constructs the H-PPO framework using triple-enhancement mechanisms (EEM, RPA, and  BCS) to realize human-collaborative endoscopic autonomous navigation. The experimental system is developed on a Unity simulation platform. To validate the method, comparative experiments include: 1) Navigation performance comparison with DRL methods (SAC, PPO, and C-PPO); 2) Component ablation studies for triple-enhancement mechanisms. Evaluation metrics cover Average Trajectory Error (ATE), Security Score (S), and Trajectory Smoothness (Jerk index), focusing on exploring the effects of exploration efficiency, safety constraint, and expert behavior imitation.
\subsection{Experimental Platform}
We aim to train an RDE reinforcement learning model to control the motion of a digestive endoscopic end-effector via decision commands output by the model. However, direct access to patients for real-world experiments is challenging. Thus, we employ a Unity-based comprehensive simulation platform \cite{incetan2021vr} to advance our research. This environment can simulate realistic tissues and organs. A monocular-camera endoscope is used for training and testing in multi-segment virtual colons. Simulations were conducted on a standard keyboard-equipped computer featuring an Intel Gold 6154 CPU and NVIDIA RTX 3090 GPU. The simulation software is Unity, with algorithms and scripts written in C\#. Neural network models were developed using the ML-Agents framework.
\begin{table}[htbp]
\centering
\caption{Human intervention commands and their corresponding keys}
\label{key_detail}
\begin{tabular}{ccc}
\toprule
\textbf{Type} & \textbf{Command} & \textbf{Key Set} \\ 
\midrule
\multirow{4}{*}{Translation} 
& Forward & Key\_up \\ 
& Backward & Key\_down \\ 
& Left Translation & Key\_left \\ 
& Right Translation & Key\_right \\ 
\addlinespace[0.3em] 
\multirow{4}{*}{Turning} 
& Upward Turn & B+Key\_up \\ 
& Downward Turn & B+Key\_down \\ 
& Left Turn & B+Key\_left \\ 
& Right Turn & B+Key\_right \\ 
\bottomrule
\end{tabular}
\end{table}

\begin{table*}[ht]
\centering
\caption{Key Evaluation Metrics}
\label{tab:metrics}
\begin{tabular}{@{}llm{6cm}@{}} 
\toprule
\textbf{Metric} & \textbf{Formula} & \textbf{Definition} \\ \midrule
Average Trajectory Error (ATE) 
& \( \text{ATE} = \frac{1}{N} \sum_{i=1}^{N} \lVert \mathbf{p}_i - \hat{\mathbf{p}}_i \rVert_2 \) 
& Mean Euclidean distance between actual (\( \mathbf{p}_i \)) and estimated (\( \hat{\mathbf{p}}_i \)) 3D points. \\ 
Jerk Index  & \( \text{Jerk} = \frac{d^3x}{dt^3} \)  
& Rate of acceleration change (\( \text{m/s}^3 \)), quantifying motion smoothness. High values indicate potential tissue damage risk. \\
Security Coefficient 
& \( S = 1 - \left( \alpha \cdot \frac{f(\mathcal{D})}{N} + \beta \cdot \frac{C}{N} + \gamma \cdot \frac{\sum F_i}{C \cdot F_{\text{thr}}} \right) \)  
& Composite safety score considering: distance violations (\( \mathcal{D} = 5\,\text{mm} \)) (Figure \ref{fig:experiment_configuration}), collision count (\( C \)), and force magnitude (\( F_i \)) with weights \( \alpha=0.4, \beta=0.5, \gamma=0.1 \). \\ 
\bottomrule
\end{tabular}
\end{table*}
\subsection{Human Guidance implementation}
We implemented human guidance to enable seamless human participation and withdrawal during reinforcement learning. Both humans and RL agents share identical perceptual information, including endoscopic images and target distance. Given that humans excel at providing heuristic commands rather than precise numerical inputs, the human action input mode is designed to facilitate heuristic guidance. In the simulation environment, humans issue commands by pressing keyboard keys. Eight commands are available: forward, backward, left, right, left turn, right turn, forward turn, and backward turn. Users indicate primary intent with a single key press (e.g., Left key), while key press duration specifies command intensity. For example, holding the Left key longer generates a larger steering value if the human is dissatisfied with the current turning degree. This converts human heuristic intent into concrete demonstration actions usable by the RL algorithm. When human actions are sent, they override the RL actions at that timestep, as detailed in Table \ref{key_detail}.

\subsection{Evaluation Metrics}
RL training was conducted in simulation environments, with overall performance evaluated using the average reward from 5 training runs. The trained RL agent was subsequently tested on the No.3 colon model. The evaluation utilized three metrics: Average Trajectory Error (ATE), Security Coefficient, and Jerk Index, as shown in Table \ref{tab:metrics}.

\subsection{Implementation Details}
\begin{figure}
    \centering
    \vfill
\includegraphics[width=1\linewidth]{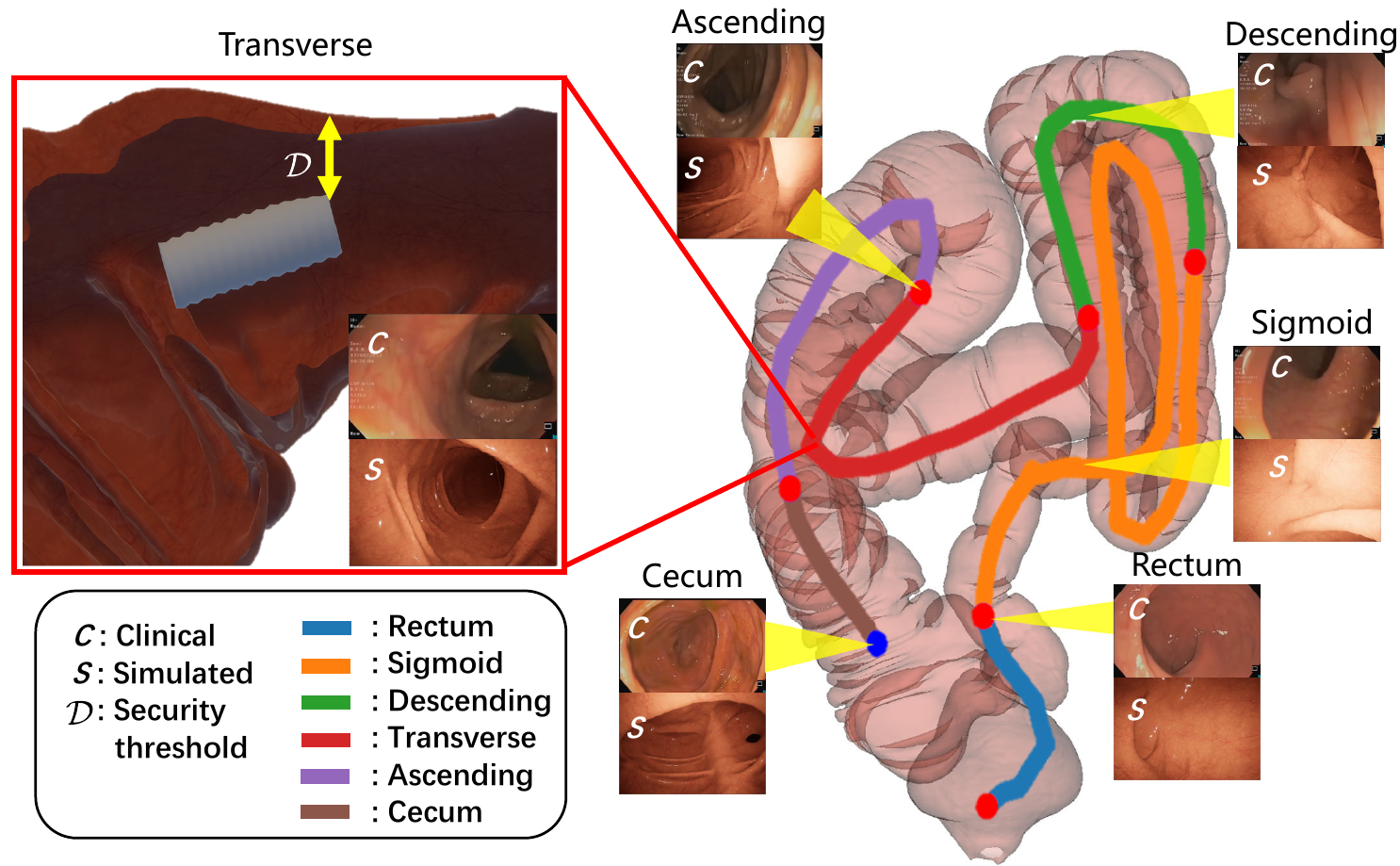}
    \caption{Experiment configuration and environmental visualization based on different anatomical segments.}
    \label{fig:experiment_configuration}
\end{figure}

\begin{figure}
    \centering
    \vfill
\includegraphics[width=1\linewidth]{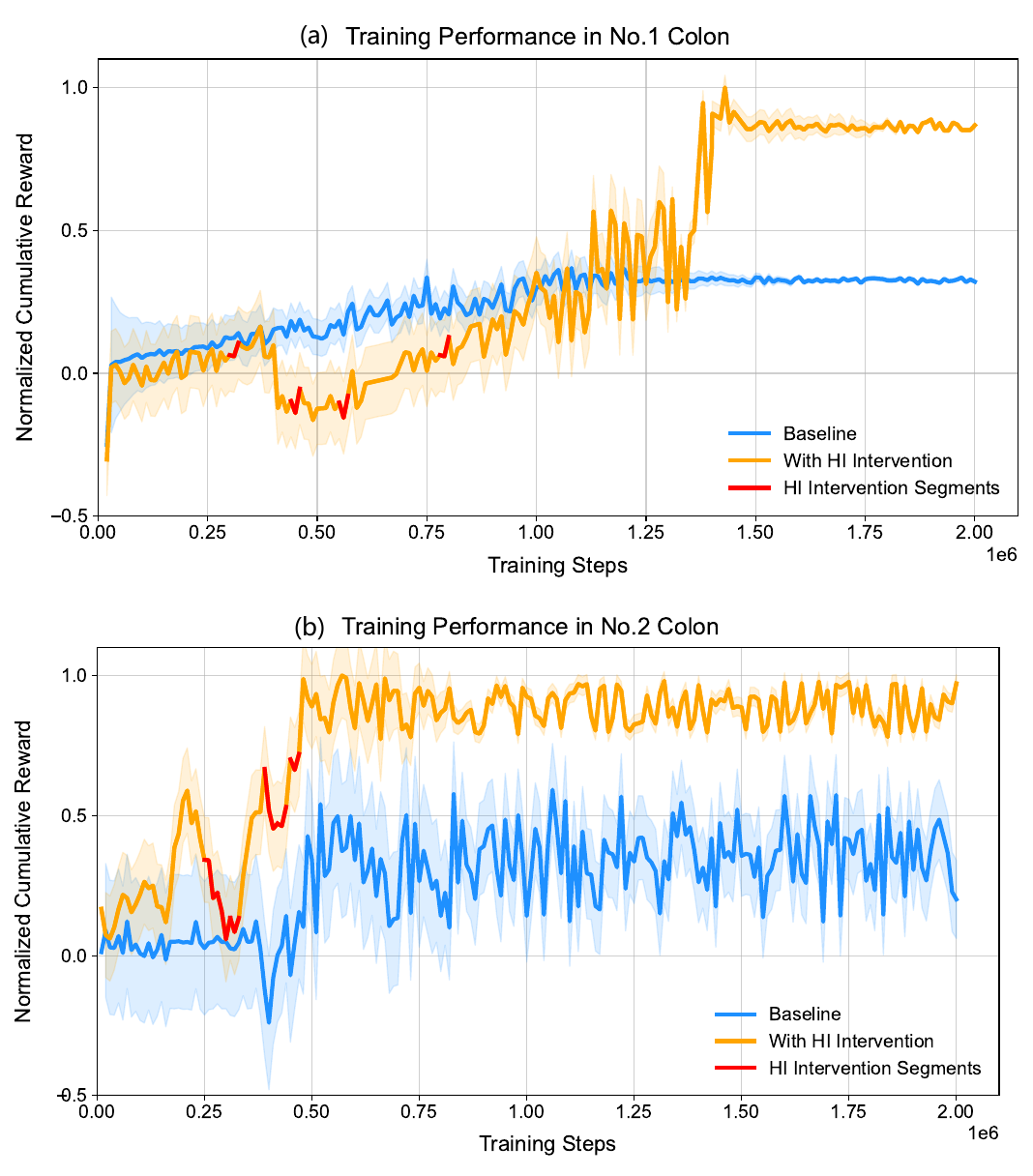}
    \caption{The learning curve of HI-PPO training on colons of different complexity. Two colon models were used. Cumulative rewards are normalized in the range [-1, 1]. The shaded area represents the range of values obtained over 5 training sessions.}
    \label{fig:trainng_curve}
\end{figure}

Training and evaluation of the proposed framework were conducted on three colon models of increasing complexity, as detailed in Figure  \ref{fig:colon_model}. The primary objective was to evaluate training differences between the proposed HI-PPO and standard PPO (Baseline) methods. The evaluation followed three sequential steps. \textbf{Step 1:} Train five instances of both algorithms for two million steps on the No.1 colon model with different random initializations. Select the top 200 policies based on success rates during training (Figure  \ref{fig:trainng_curve}). \textbf{Step 2:} Continue training these 200 policies on the No.2 colon model, retaining the best ten models. \textbf{Step 3:} Test the final models on the No.3 colon model, with segment-specific evaluation according to difficulty levels (Figure  \ref{fig:experiment_configuration}).

\section{Results}
\subsection{Impact of Human Intervention on Training}
First, we evaluated and compared the proposed HI-PPO algorithm with the Baseline. A fair comparison was achieved by training both methods five times using the same random seed sequence. Humans intervened only in the early training phase for the human-involved HI-PPO algorithm. The rewards during training are shown in Figure  \ref{fig:trainng_curve}. In the No.1 colon, the "With HI Intervention" group (orange) saw a significant increase in normalized cumulative reward as training steps progressed, far exceeding the Baseline (blue) in later stages, with minimal fluctuation across 5 training runs. In the No.2 colon, the HI group's rewards fluctuated significantly, still outperforming the Baseline. The red segments in the figure represent the human intervention process. Results show that the Baseline's curve rose slowly due to the lack of an effective exploration mechanism. The HI intervention mechanism aids model learning, leading to more remarkable cumulative reward improvements. These results validate the advantage of HI-PPO in optimizing the learning process and enhancing training performance through intervention in colon models of varying complexities, proving more effective than the baseline method.

\begin{table*}[h]
\centering
\caption{Quantitative Comparison Across Anatomical Segments (ATE: mm, Security Coefficient: \(S\))}
\label{tab:ATE_result}
\scalebox{0.8}{
\begin{tabular}{lcccccccccc}
\toprule
\multirow{2}{*}{\textbf{Colon Segment}} & \multicolumn{2}{c}{\textbf{SAC}} & \multicolumn{2}{c}{\textbf{PPO}} & \multicolumn{2}{c}{\textbf{C-PPO}} & \multicolumn{2}{c}{\textbf{HI-PPO}} & \multicolumn{2}{c}{\textbf{Human Expert}} \\
\cmidrule(lr){2-3}\cmidrule(lr){4-5}\cmidrule(lr){6-7}\cmidrule(lr){8-9}\cmidrule(lr){10-11}
 & ATE (Std) & Security & ATE (Std) & Security & ATE (Std) & Security & ATE (Std) & Security & ATE (Std) & Security \\
\midrule
\textbf{Rectum} 
& 4.915 (0.977) & 0.5531 
& 4.687 (0.590) & 0.5867 
& 3.523 (0.856) & 0.6478 
& \underline{2.353 (0.629)} & \underline{0.7749} 
& \textbf{1.98 (0.48)} & \textbf{0.8121} \\

\textbf{Sigmoid} 
& 20.12 (13.61) & 0.5785 
& 33.38 (17.88) & 0.6286 
& 22.15 (12.43) & 0.6964 
& \textbf{16.15 (11.37)} & \textbf{0.8314} 
& \underline{17.12 (10.82)} & \underline{0.8303} \\

\textbf{Descending}
& 20.28 (3.882) & 0.6853 
& 17.97 (7.271) & 0.7438 
& 13.82 (6.154) & 0.7687 
& \underline{9.269 (3.687)} & \underline{0.8465} 
& \textbf{8.99 (3.05)} & \textbf{0.8590} \\

\textbf{Transverse}
& 15.35 (9.062) & 0.7850 
& 9.001 (6.616) & 0.8140 
& 8.673 (5.824) & 0.8659 
& \textbf{8.348 (5.699)} & \underline{0.8970} 
& \underline{8.77 (5.41)} & \textbf{0.9021} \\

\textbf{Ascending} 
& 9.53 (5.659) & 0.7789 
& 15.80 (7.616) & 0.8274 
& 12.25 (6.832) & 0.8574 
& \underline{8.436 (4.004)} & \underline{0.9145} 
& \textbf{7.85 (3.65)} & \textbf{0.9202} \\

\textbf{Cecum} 
& 12.20 (2.199) & 0.7888 
& 3.984 (2.329) & 0.7927 
& 3.762 (2.058) & 0.8638 
& \underline{3.597 (1.929)} & \underline{0.9107} 
& \textbf{3.49 (1.85)} & \textbf{0.9874} \\
\midrule
\textbf{Mean (Approx) } 
& 13.73  & 0.695 
& 14.14 & 0.732
& 10.70 & 0.783 
& \textbf{8.02} &  \underline{0.862 }
& \underline{8.03} &\textbf{0.885} \\
\bottomrule
\end{tabular}
}
\end{table*}    

\begin{figure*}
    \centering
    \vfill
\includegraphics[width=0.8\linewidth]{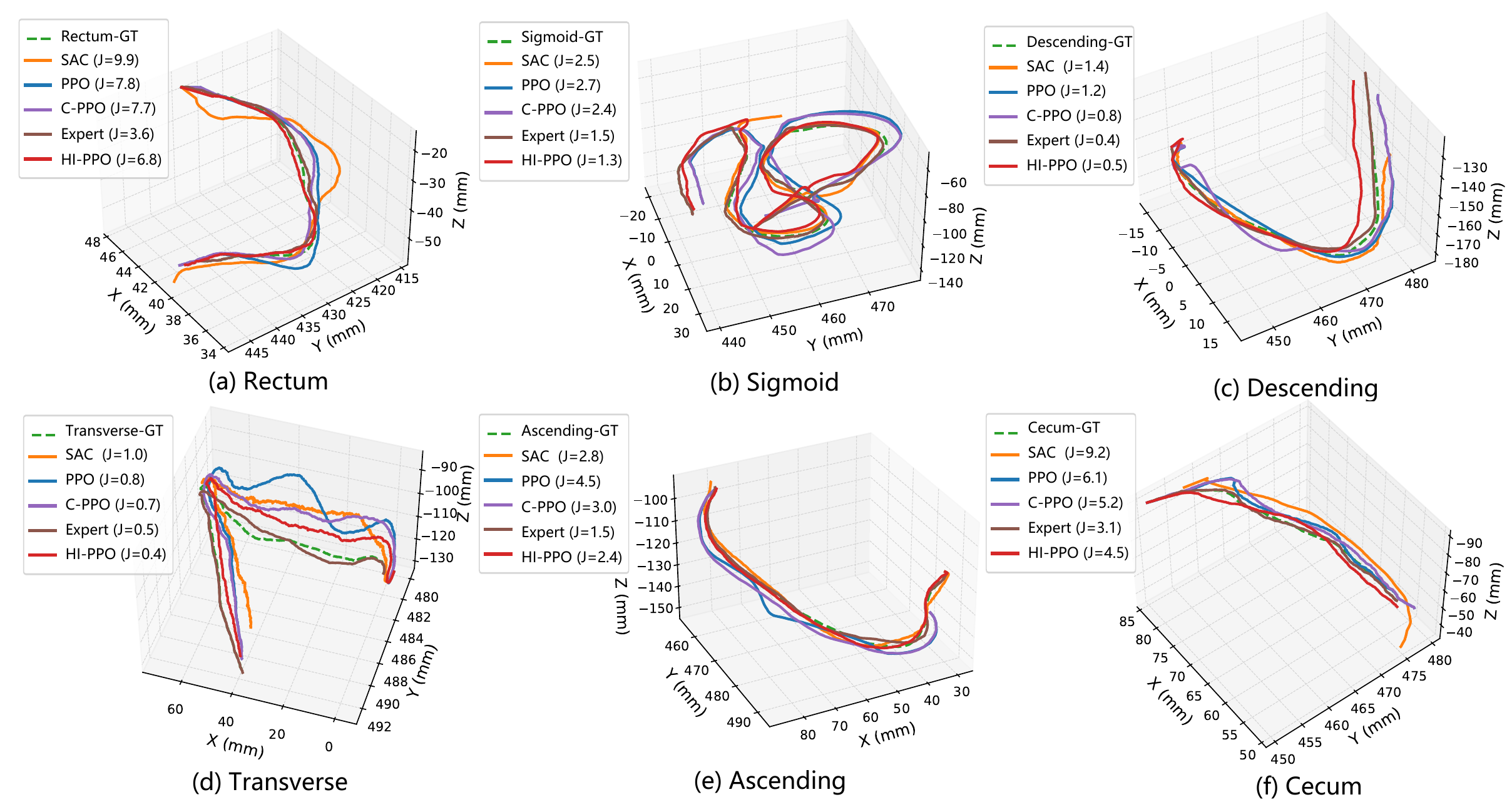}
    \caption{Visualization of navigation trajectories in different anatomical segments. Jerk index (J, unit: mm/s³) is overlaid to quantify motion smoothness, where lower values indicate reduced mechanical stress on the intestinal mucosa.}
    \label{fig:res_traj}
\end{figure*}
\subsection{Quantitative Comparison and Evaluation}
Then, we validated the effectiveness of HI-PPO components in the No.3 colon model using a segmented evaluation strategy. For anatomical differences, such as the rectum’s complex bending, the ascending colon’s entanglement tendency, and the cecum’s dilated cavity-phased performance quantification identifies each module’s performance bottlenecks under specific anatomical forms, providing an anatomical basis for optimizing the navigation algorithm. We compared our method against three baselines. SAC \cite{haarnoja2018soft}, a widely used off-policy reinforcement learning algorithm, maximizes rewards and entropy while promoting diverse exploration. The standard PPO \cite{schulman2017proximal} without human intervention served as a baseline, and C-PPO, a PPO with strong constraints, followed \cite{corsi_constrained_2023}. All methods underwent parameter-tuning training, with the ground-truth (GT) path as the colon’s central line.
Compared to the three baselines, our method consistently achieved lower ATE in colon segments of varying lengths and complexities (Table \ref{tab:ATE_result}). Specifically, HI-PPO reduced the average tracking error (ATE) by 3.7\%–33.2\% versus suboptimal algorithms across all 6 segments. The descending colon’s ATE was 9.269 mm, only 3.01\% behind human experts. Standard deviation results show our method’s higher stability and minimal error fluctuation. Clinically, with the colon’s average diameter at ~70 mm \cite{bassler1920diseases}, a 10 mm margin is a safe distance in robot-assisted colon surgery \cite{park2016robotic}\cite{marks2024safety}. Our method stayed within this range except in the complex sigmoid segment. The sigmoid’s safety coefficient \(S = 0.8314\) exceeded human experts, verifying the obstacle-avoidance ability of our method in complex bends. The human intervention module boosted cecum security (0.9107) by 5.4\% compared to C-PPO. HI-PPO’s 8.436 mm ATE set a record in the ascending colon, though 6.9\% behind the human’s 7.85 mm due to path drift from the lengthy tract. In addition, HI-PPO outperformed baselines in safety across segments: the rectum’s \(S = 0.7749\) was 19.6\% higher than C-PPO. While the sigmoid’s ATE (16.15 mm) slightly bested human experts (17.12 mm), a standard deviation of 11.37 mm indicates stability needs further improvement. Overall, HI-PPO balances safety and accuracy, laying a technical foundation for autonomous colonoscopy navigation.
\begin{table*}[h]
\centering
\caption{Ablation Study of HI-PPO Across Anatomical Segments (ATE: mm)}
\label{tab:ablation}
\scalebox{0.82}{
\begin{tabular}{lccccccccc}
\toprule
\textbf{Colon Segment} & \textbf{PPO} & \textbf{+EEM} & \textbf{+RPA} & \textbf{+BCS} & \textbf{EEM+} & \textbf{EEM+} & \textbf{RPA+} & \textbf{HI-PPO} \\
& \textbf{(Baseline)} & & & & \textbf{RPA} & \textbf{BCS} & \textbf{BCS} & \textbf{(Full)} \\
\midrule
\textbf{Rectum}    & 4.687 (0.590) & 3.952 (0.621) & 4.123 (0.602) & 3.845 (0.598) & 3.521 (0.615) & 3.312 (0.607) & 3.674 (0.594) & \textbf{2.353 (0.629)} \\
\textbf{Sigmoid}   & 33.38 (17.88) & 28.15 (15.23) & 29.74 (16.02) & 27.89 (14.97) & 24.57 (13.45) & 22.83 (12.69) & 25.41 (13.87) & \textbf{16.15 (11.37)} \\
\textbf{Descending}& 17.97 (7.271) & 14.32 (6.854) & 15.08 (7.123) & 13.75 (6.215) & 12.04 (5.987) & 11.29 (5.432) & 12.87 (6.102) & \textbf{9.269 (3.687)} \\
\textbf{Transverse}& 9.001 (6.616) & 8.892 (5.823) & 8.945 (6.102) & 8.654 (5.632) & 8.814 (5.712) & 8.645 (5.987) & 8.892 (5.614) & \textbf{8.348 (5.699)} \\
\textbf{Ascending} & 15.80 (7.616) & 12.45 (6.324) & 13.27 (6.785) & 11.98 (6.012) & 10.23 (5.432) & 9.854 (5.112) & 10.67 (5.687) & \textbf{8.436 (4.004)} \\
\textbf{Cecum}     & 3.984 (2.329) & 3.852 (2.015) & 3.867 (2.104) & 3.821 (1.987) & 3.612 (1.954) & 3.625 (1.992) & 3.614 (1.945) & \textbf{3.597 (1.929)} \\
\bottomrule
\end{tabular}
}
\end{table*}

\subsection{Ablation of Human Intervention Module}
As shown in Table \ref{tab:ablation}, the HI-PPO method demonstrates significant advantages in colon navigation tasks. The complete model reduces the average tracking error (ATE) by 49.8\% and 51.6\% compared to baseline PPO in key anatomical segments like the rectum and sigmoid colon. In the complex sigmoid colon, the collaboration of the EEM environment-exploration module and RPA path-planning cuts errors from 33.38 mm to 24.57 mm, with standard deviation shrinking by 24.7\%, showcasing these two modules' adaptability to complex anatomy. BCS plays a key role in the descending colon, enabling the HI-PPO model to achieve the 13.75 mm, resulting in a 23.5\% improvement over single modules. The cecum's error of HI-PPO (3.597 mm) needs improvement, likely due to path cumulative effects. A backtracking correction mechanism is planned for improvement.

\subsection{Navigation Trajectory Evaluation}
 From Figure \ref{fig:res_traj}, HI-PPO provides a more reliable movement strategy, approximating expert operations for colon navigation tasks. Though the ATE in the sigmoid colon segment is prominent due to its length and curvature, our method maintains trajectory consistency with GT. As shown in Figure  \ref{fig:res_traj}, compared with SAC (J = 9.9) and PPO (J = 7.8), HI-PPO reduces deviations in complex segments like the rectum (J = 6.8) and transverse colon (J = 0.4), verifying its ability in motion smoothness. In the sigmoid colon, the SAC (J = 2.5) shows apparent deviations in the latter part of turns, deviating from the expected path in the intestinal environment. HI-PPO effectively alleviates the severe oscillation of other RL algorithms by integrating human intervention skills. It is also can be observed that our method can actively adjust the movement direction during sharp turns to reduce acceleration peaks (Figure  \ref{fig:res_traj} (b)). In the descending colon Figure \ref{fig:res_traj} (c) and transverse colon Figure \ref{fig:res_traj} (d), HI-PPO’s J values (0.5 and 0.4) remain low, with trajectories closer to the expert path than algorithms like SAC and PPO. In the ascending colon Figure  \ref{fig:res_traj} (e) and cecum Figure  \ref{fig:res_traj} (f), the Jerk index of our method is not the lowest compared to the expert. Overall, HI-PPO improves trajectory stability compared with SAC, PPO, and C-PPO, decreasing stress concentration on intestinal mucosa via safety mechanisms. It achieves smoother motion control in most colon segments, validating its effectiveness in navigation trajectory optimization.

\begin{figure}[tbp]
    \centering
    \hfill
    \includegraphics[width=1\linewidth]{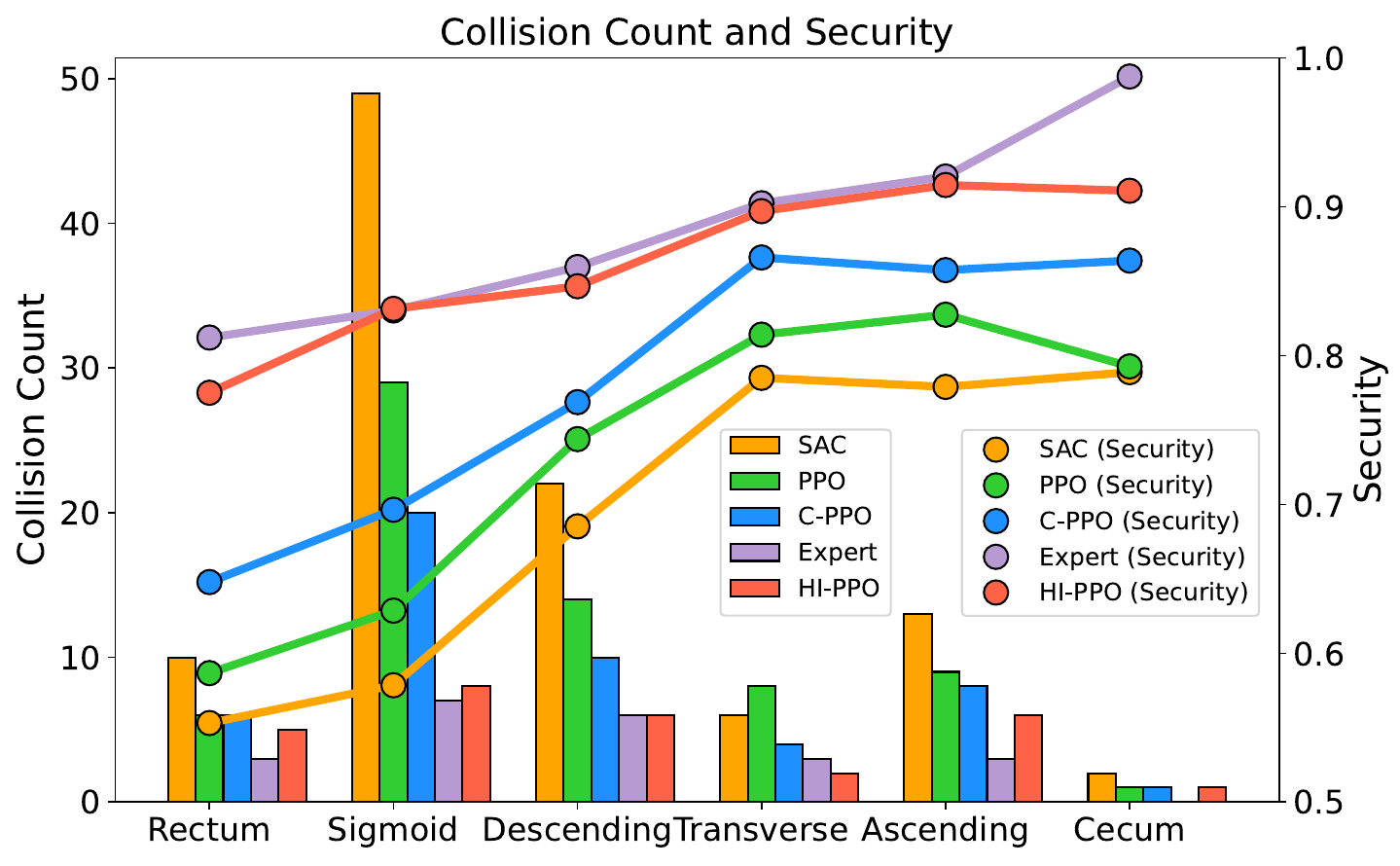}
    \caption{Collision and security}
    \label{fig:collision_show}
\end{figure}
\begin{figure}[tbp]
    \centering
    \hfill
    \includegraphics[width=1\linewidth]{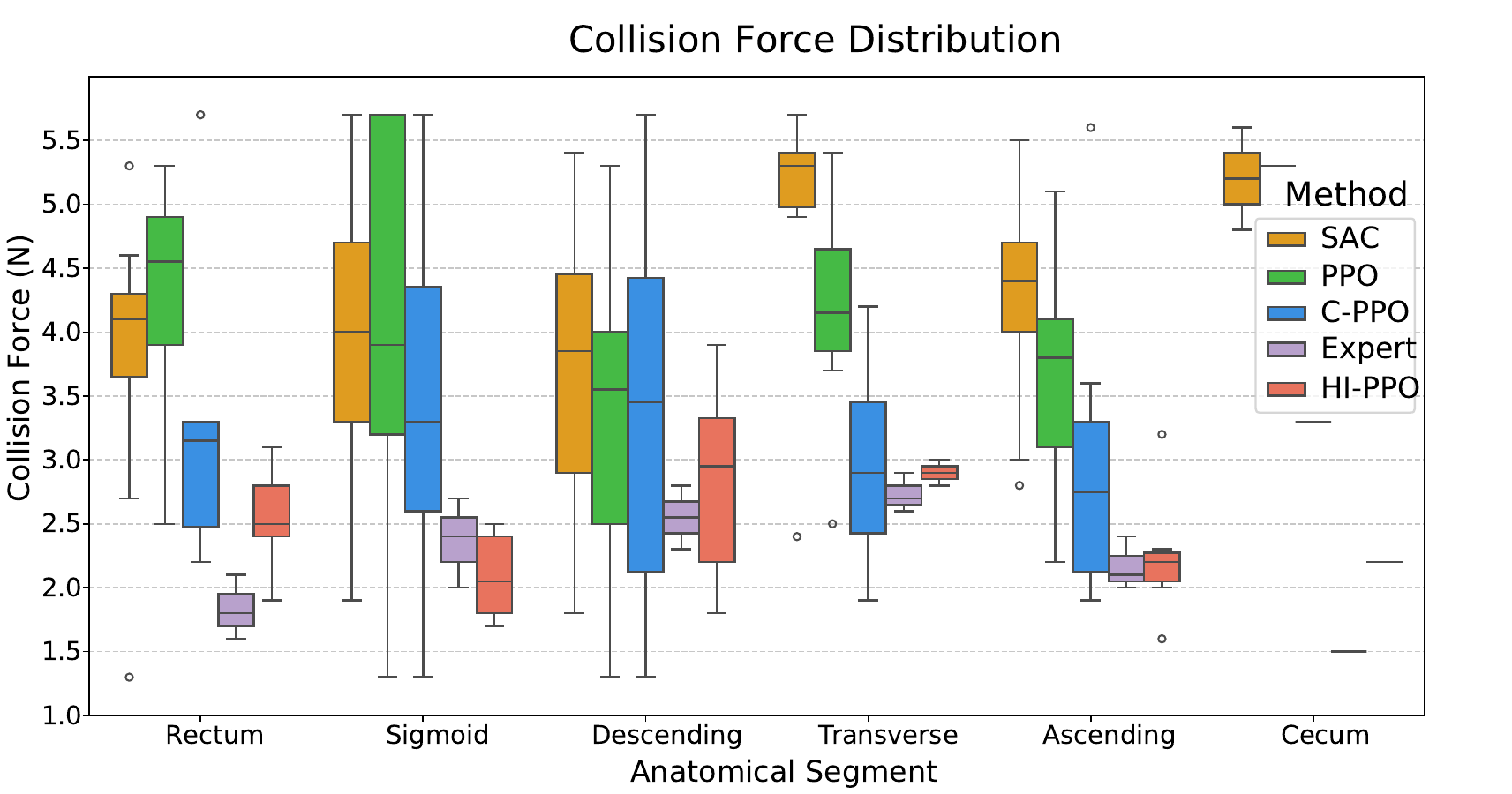}
    \caption{Collision Force Distribution}
    \label{fig:collision_force}
\end{figure}

\subsection{Safety Evaluation}
Figure \ref{fig:collision_show} presents collision counts and safety coefficients of different models across colon segments. In the rectum, HI-PPO’s collision count (10 collisions) is lower than SAC (10 collisions) and PPO (6 collisions), with a safety coefficient of 0.77, which is higher than SAC’s 0.55 and PPO’s 0.58. In the sigmoid colon, HI-PPO’s collision count (8 collisions) is lower than PPO’s 29 collisions, and its safety coefficient exceeds 0.83, notably better than SAC’s 0.57. In descending and transverse colon regions, HI-PPO maintains low collision counts and stable safety coefficients of $\geq$0.84. Compared to experts, HI-PPO’s safety coefficients in multiple segments closely match expert levels (e.g., transverse and ascending). Overall, HI-PPO achieves lower collision counts (e.g., rectum and transverse) and higher safety coefficients in most segments than SAC and PPO through human intervention techniques, effectively balancing collision control and safety. This provides a more reliable safety strategy for colon navigation scenarios.

Figure \ref{fig:collision_force} shows collision force distributions of different methods across colon segments. HI-PPO’s collision forces in the rectum concentrate at 2.0 $\sim$ 2.8 N, lower than SAC (3.5$\sim$4.5 N) and PPO (3.5$\sim$5.0 N). HI-PPO’s force quartile (1.7$\sim$2.5 N) is also lower than other methods in the sigmoid colon. In descending and transverse colon regions, HI-PPO’s median collision forces remain lowest (e.g., transverse colon 2.7 N vs. C-PPO’s 2.9  N). These results benefit from HI-PPO’s security mechanism integration detailed in Section 3.3\ref{Securitymechanismintegration} that regulates operational forces and reduces excessive intestinal wall contact. Compared to SAC and PPO, HI-PPO achieves superior collision force control across all segments, reducing mechanical damage risks to the intestinal mucosa. This validates its safety advantages from a mechanical interaction perspective, offering a more reliable force control solution for colon navigation.
\section{Discussion}
The proposed HI-PPO framework represents an advancement in robotic digestive endoscopy by integrating human intervention into reinforcement learning to address potential safety challenges in complex anatomical environments. However, several limitations warrant consideration. First, our current safety evaluation focuses solely on the endoscope tip, neglecting interactions between the flexible shaft and intestinal walls. Future work will expand the model to incorporate full endoscope dynamics. Second, while the Unity simulation faithfully reconstructs colon morphology, it lacks real-time tissue deformation, which may affect force feedback accuracy. Addressing this requires integrating finite element models of colon mechanics into the simulation framework.  

In the future, we will conduct ex-vivo model validation experiments, aiming to bridge the simulation-to-reality gap and demonstrate the advanced optimization in balancing motion accuracy and tissue protection. Additionally, exploring hierarchical policy architectures, where high-level planning handles segment transitions and low-level control manages local navigation, could improve adaptability in tortuous pathways. Clinically, the framework allows for seamless integration with commercial endoscopic systems, positioning it as a scalable solution for enhancing procedural safety.

\section{Conclusions}
This study addresses the challenges of safe and efficient navigation in automated robotic digestive endoscopy (RDE) within the unstructured and narrow confines of the digestive tract. We proposed a human intervention-based PPO framework by incorporating expert knowledge to address the safety and effectiveness of RDE. Experiments have shown that HI-PPO can safely guide RDE compared to existing RL algorithms, indicating its potential for more practical application. Future work will focus on validating the framework in real environments and exploring additional methods to further enhance its safety and practicality.
\section*{Acknowledgements}
We sincerely appreciate the support from the doctors in the Department of Gastroenterology and Hepatology at Shenzhen University General Hospital.

\bibliographystyle{unsrt}  






\end{document}